\documentclass{svproc}

\usepackage{graphicx}
\usepackage{color,soul}
\usepackage{lmodern} 
\usepackage{amsmath} 
\usepackage{amssymb} 
\usepackage{epigraph}

\usepackage{url}

\newcommand{\R}{\mathbb{R}}
\newcommand{\I}{\mathbb{I}}

\begin{document}
\mainmatter              

\title{A Computational-Hermeneutic Approach \\
	for Conceptual Explicitation\thanks{Benzm\"uller received
          funding for this research from VolkswagenStiftung under grant CRAP 93678 (Consistent Rational Argumentation in Politics).}}
\titlerunning{A Computational-Hermeneutic Approach}  
\author{David Fuenmayor\inst{1} \and Christoph Benzm\"uller\inst{1,2}}
\authorrunning{D. Fuenmayor and C. Benzm\"uller} 

%
\institute{Freie Universit\"at Berlin, Germany
	\and
	University of Luxembourg, Luxembourg
}

\maketitle              

\begin{abstract}
We present a computer-supported approach for the logical analysis and conceptual explicitation of argumentative discourse.
Computational hermeneutics harnesses recent progresses in automated reasoning for higher-order logics and aims at formalizing natural-language argumentative  discourse using flexible combinations of expressive non-classical logics.
In doing so, it allows us to render explicit the tacit conceptualizations implicit in argumentative discursive practices.
Our approach operates on networks of structured arguments and is iterative and two-layered.
At one layer we search for logically correct formalizations for each of the individual arguments.
At the next layer we select among those correct formalizations the ones which
honor the argument's dialectic role, i.e. attacking or supporting other arguments as intended.
We operate at these two layers in parallel and continuously rate sentences' formalizations
by using, primarily, inferential adequacy criteria. An interpretive, logical theory will thus
gradually evolve. This theory is composed of meaning postulates serving as explications for concepts playing a role in the analyzed arguments.
Such a recursive, iterative approach to interpretation does justice to the inherent circularity of understanding: the whole is understood compositionally on the basis of its parts,
while each part is understood only in the context of the whole (hermeneutic circle).
We summarily discuss previous work on exemplary applications of human-in-the-loop computational hermeneutics in metaphysical discourse. We also discuss some 
of the main challenges involved in fully-automating our approach. 
By sketching some design ideas and reviewing relevant technologies, we argue for the technological feasibility of a highly-automated computational hermeneutics.

\end{abstract}
\setlength\epigraphwidth{.8\textwidth}
\epigraph{``\dots that the same way that the whole is, of course, understood
	  		      in reference to the individual, so too, the individual can
  		      only be understood in reference to the whole.''}{Friedrich Schleiermacher (1829)}

\newpage
\section{Introduction}
Motivated by, and reflecting upon, previous work on the
computer-supported assessment of challenging arguments in metaphysics
(e.g. \cite{C55,J32,C65}), we have engaged in the development of a systematic approach towards the logical
analysis of natural-language arguments amenable to (partial)
automation with modern theorem proving 
technology. 
In previous work \cite{J38,B18} we have presented some case studies
illustrating a computer-supported approach, termed
\textit{computational hermeneutics}, which has the virtue of
addressing argument formalization and assessment in a holistic way: The adequacy of candidate formalizations for sentences in an argument is assessed by computing the logical validity of the argument as a whole (which depends itself on the way we have so far formalized all of its constituent sentences).
Computational hermeneutics has been inspired by ideas in the philosophy of language such as \textit{semantic holism} and Donald Davidson's \textit{radical interpretation} \cite{davidson2001inquiries}. It is grounded on recent progresses in the area of automated reasoning for higher-order logics and  integrates techniques from argumentation theory \cite{B18}.

Drawing on the observation that ``every formalization is an
interpretation'', our approach has initially aimed at fostering
human understanding of argumentative (particularly theological) discourse \cite{J38,C74}.
In the following sections, aside from presenting a more detailed theoretical account of our approach, we want to explore
possibilities for further automation, thus gradually removing the human from the loop.		
In section \ref{sec:conceptualizations} we introduce the notions of \textit{ontology} and \textit{conceptualization}
as defined in the fields of knowledge engineering and artificial intelligence.
We use this terminological framework to discuss what it means for us
to claim that a logical theory can serve to \textit{explicitate a conceptualization}.
In section \ref{sec:hermeneutic} we discuss what makes our proposed approach hermeneutical.
We briefly present a modified account of Donald Davidson's theory of radical interpretation and relate it to the
logical analysis of natural-language arguments, in particular, to the problem of assessing the adequacy of sentences'
formalizations in the context of an argument (network). We show how an interpretive process grounded on the logical
analysis of natural-language argumentative discourse will exhibit a
(virtuous) circularity, and thus needs to be approached in a mixed
recursive--iterative way.

We also illustrate the fundamental role of computers in supporting this process.
In particular, by automating the computation of inferential adequacy criteria of formalization,
modern theorem proving technology for higher-order logics can provide the
effective and reliable feedback needed to make this approach a viable alternative.
In section \ref{sec:implementation} we briefly present an exemplary case study
involving the analysis of a metaphysical argument and discuss some technologies and
implementation approaches addressing some of the main challenges 
concerning a highly-automated computational hermeneutics. In particular, we discuss a key technique termed \textit{semantical embeddings}~\cite{J23,J41}, which
harnesses the expressive power of classical higher-order logic to enable a truly logical-pluralist approach towards representing
and reasoning with complex theories by reusing state-of-the-art (higher-order) automated reasoning infrastructure.

\section{Explicitating Conceptualizations} \label{sec:conceptualizations}

Computational hermeneutics helps us make our \textit{tacit} conceptualizations \textit{explicit}.
In this section we aim at building the necessary background to make best sense of this statement.
We start with a cursory presentation of the notion of \textit{an} ontology as used in the fields of artificial intelligence and knowledge engineering. We introduce some of the definitions of (an) ontology drawing on the concept of conceptualization
(this latter notion being the one we are mostly interested in). We then have a deeper look into the notion of conceptualizations
and how they can be explicitly represented by means of a logical theory.

\subsection{Ontologies and Meaning Postulates}  \label{subsec:conceptualizations-ontologies}

Ontology is a quite overloaded term, not only in its original philosophical scope, but also in computer science.
Trying to improve this situation, researchers have made the apt distinction between ``Ontology'' and ``an ontology'' \cite{guarino1995ontologies}.
The term ``Ontology'' refers to the philosophical field of study, which will not be considered in this paper.
Regarding the latter term (``an ontology'') several authoritative definitions have been put forward during the nineties. We recall some of them:
Tom Gruber originally defines an ontology as ``an explicit specification of a conceptualization'' \cite{gruber1993translation}.
This definition is further elaborated by Studer et al.~\cite{studer1998knowledge} as ``a formal, explicit specification of a shared conceptualization'', 
thus emphasizing the dimensions of intersubjective conspicuousness and representability by means of a formal language.
Nicola Guarino, a pioneer advocate of ontologies in computer science, has depicted an ontology as
``a logical theory which gives an explicit, partial account of a conceptualization'' \cite{guarino1995ontologies},
thus emphasizing an aspect of insufficiency: an ontology can only give a \textit{partial} account of a conceptualization.

These definitions duly highlight the aspect of explicitness: By means of the articulation in a formal language,
an ontology can become common and conspicuous enough to fulfill its normative role ---as public standard---
for the kind of systems developed in areas like knowledge engineering and, more recently, the Semantic Web.
More technically, we see that an ontology (at least in computer science) can be aptly considered as a kind of logical theory,
i.e. as a set of sentences or formulas. The question thus arises: Which particular formal properties should the sentences
comprising a logical theory have in order to count as an ontology?
To be sure, the logical theory may feature some particular annotation or indexing schema distinguishing `ontological' sentences from
`non-ontological' ones. But other than this there is no clear line outlining the `ontological' sentences in a logical theory.\footnote{
	This is similar to the distinction between \textit{TBox} and \textit{ABox} in knowledge bases.
	Some may claim that `ontological' sentences (TBox) tend to be more permanent and mostly concern types, classes and other universals;
	while other, `non-ontological' sentences (ABox) mostly concern their instances.
	The former may be treated as being always true and the latter as subject to on-line revision.
	However, what counts as a class, what as an instance and what is subject to revision is heavily
	dependent on the use we intend to give to the theory (knowledge base).}

This issue is reminiscent of the controversy around the old analytic--synthetic distinction in philosophy.
We do not want to address this complex debate here. However, we want to draw attention to a related duality introduced by
Carnap \cite{carnap1952meaning} in his notion of \textit{meaning postulates} (and its counterpart: \textit{empirical postulates}).
Carnap's position is that, in order to rigorously specify a language ---or a logical theory, for our purposes--- one is confronted with a
\textit{decision} concerning which sentences are to be taken as analytic, i.e. which ones we should attach the label ``meaning postulate''.
For Carnap, meaning postulates are axioms of a definitional nature, which tell us how the meanings
of terms are interrelated.\footnote{Recalling Carnap's related notion of \textit{explication} \cite{carnap1947meaning}, 
	we can think of a set of meaning postulates as providing a precise characterization for some new, exact
	concept (\textit{explicatum}) aimed at ``replacing'' an inexact, pre-theoretical notion (\textit{explicandum}),
	for the purpose of advancing a theory.
	Thus, in computational hermeneutics, the non-logical terms of our interpretive theory
	characterize concepts playing the role of \textit{explicata} aimed at explicitly representing fuzzy, unarticulated \textit{explicanda} from a tacit conceptualization.	
}
They are warranted by an \textit{intent to use} those terms in a certain way (e.g. to draw some inferences that we accept as valid)
rather than by any appeal to facts or observations.
In this sense, meaning postulates are to be distinguished from factual assertions,
also termed ``empirical postulates''. A kind of analytic--synthetic distinction is thus made, but this time by an appeal to pragmatical considerations.
Whether a sentence such as ``No mammals live in water'' is analytic, or not, depends upon a decision
about how to \textit{use} the corresponding concepts in a certain area of discourse; e.g. we may want to favor some
inferences and disfavor others (e.g. excluding whales from being considered as mammals).
We thus start to see how a listing of meaning postulates can give us some insight into the conceptualization
underlying a discourse. Following Carnap, we distinguish in our approach between meaning and empirical postulates.
The former are sets of axioms telling us how the meanings of terms are interrelated, and thus 
constitute our interpretive logical theory.
The latter correspond to formalized premises and conclusions of arguments which are to become
validated in the context of our theory.

As discussed above, it is rather the intended purpose which lets us consider a logical theory (or a part thereof) as an ontology.
We will thus conflate talk of formal ontologies with talk of logical theories for the sake of illustrating our
computational-hermeneutic approach.

\subsection{Conceptualizations} \label{subsec:conceptualizations-2}

As seen from the above, definitions of (an) ontology depict a conceptualization, by contrast, as something tacit and unarticulated;
as something being \textit{hinted at} by the use of natural language, instead of being \textit{determined by} the semantics of a formal one. In the following, we want to evoke this connotation every time we use the term conceptualization \textit{simpliciter}.
Our notion of conceptualization (\textit{simpliciter}) thus refers to that kind of implicit, unarticulated and
to some extent undetermined knowledge being framed by human socio-linguistic practices,
and which is to become (partially) represented ---and thereby \textit{made explicit}--- by means of a logical theory.
In particular, we envisage a certain special ---and arguably primordial--- kind of socio-linguistic practice: argumentative discourse.

We have a formal definition for the notion of conceptualization starting with Genesereth and Nilsson \cite{genesereth1987logical},
who state that ``the formalization of knowledge in declarative form begins with a \textit{conceptualization}.
This includes the objects presumed or hypothesized to exist in the world and their interrelationships.''
In their account, objects ``can be anything about which we want to say something.''
Genesereth and Nilsson then proceed to formally define a conceptualization as an \textit{extensional} relational structure:
a tuple ${\langle}D, R{\rangle}$ where $D$ is a set called the universe of discourse and $R$ is a collection of relations (as sets of tuples) on $D$.
This account of a conceptualization has been called into question because of its restricted extensional nature:
A conceptualization, so conceived, concerns only how things in the universe of discourse \textit{actually are}
and not how they \textit{could acceptably be} interrelated. The structure proposed by Genesereth and Nilsson as a formal characterization of the notion of conceptualization seems to be more appropriate for states of affairs or world-states;
a conceptualization having a much more complex (social or mental) nature.

Other, more expressive definitions for the notion of conceptualization have followed.
Gruber \cite{gruber1993translation} states that a conceptualization is
``an abstract, simplified view of the world that we wish to represent for some purpose''; and Uschold \cite{uschold1996building} sees a conceptualization as a
``world-view'', since it ``corresponds to a way of thinking about some domain''.
According to the account provided by Guarino and his colleagues \cite{guarino1995ontologies,guarino2009ontology},
a conceptualization ``encodes the implicit rules constraining the structure of a piece of reality''.
Note that this characterization, in contrast to the one provided by Genesereth and Nilsson,
does not talk about how a ``piece of reality'' actually is, but instead about how it can possibly or acceptably be,
according to some constraints set by \textit{implicit} rules.\footnote{
	As we see it, those rules are (their tacitness notwithstanding) of a logical nature:
	they concern which arguments or inferences are endorsed by (a community of) speakers.}
Guarino has also provided a formal account of conceptualizations, which we reproduce below (taken from \cite{guarino2009ontology}).
In the following, we distinguish the informal notion of conceptualization \textit{simpliciter}
from the formal, exact concept introduced below.
Note that we will always refer to the latter in a qualified form as a \textit{formal} conceptualization.

\begin{definition}[Intensional relational structure, or formal conceptualization]
	An intensional relational structure (or a conceptualization according to Guarino) is a triple $C = {\langle}D,W,\R{\rangle}$ with
	\begin{itemize}
	\item{D} a universe of discourse, i.e. an arbitrary set of individual entities.
	\item{W} a set of possible worlds (or world-states). Each world is a maximal observable state of affairs, i.e. a
	unique assignment of values to all the observable variables that characterize the system.
	\item{$\R$} a set of intensional (aka. conceptual) relations.
	An intensional relation is a function mapping possible worlds to extensional relations on $D$.
	\end{itemize}
\end{definition}
\noindent{Taking inspiration from Carnap \cite{carnap1947meaning} and Montague \cite{dowty2012introduction},
Guarino takes conceptual (intensional) relations to be functions from worlds to extensional relations (sets of tuples).
This account aims at doing justice to the intuition that conceptualizations ---being arguably about concepts---
should not change when some particular objects in the world become otherwise related.
However, \textit{formal conceptualizations}, while having the virtue of being exactly defined and unambiguous,
are just fictional objects useful to clarify the notion of an ontology.}

\subsection{Representing a Conceptualization by Means of a Theory} \label{subsec:conceptualizations-representation}

For Guarino and his colleagues, the role of formal conceptualizations in the definition of an ontology
is that of a touchstone. For them an ontology is ---let us recall their definition---
``a logical theory which gives an explicit, partial account of a conceptualization''
\cite{guarino1995ontologies}.
Thus, an ontology ---or more specifically: its models\footnote{
	When we talk of models of a logical theory or ontology, we always refer to models in a model-theoretical sense,
	i.e. interpretations: assignments of values (as denoted entities) to non-logical terms.}--- can only \textit{partially} represent a conceptualization (formal or \textit{simpliciter}).
In order to evaluate how good a given ontology represents a conceptualization, Guarino has introduced
a formal notion of \textit{ontological commitment} (similar in spirit to the well-known eponymous notion in philosophy).
According to Guarino, by using a certain language a speaker commits ---even unknowingly---
to a certain conceptualization. Such a commitment arises from the fact that, by employing a certain
linguistic expression (e.g. a name or an adjective), a language user \textit{intends} to refer to objects
which are part of a conceptualization, i.e. some individual or (conceptual) relation.
Those objects may be tacit in the sense that they become presupposed in our inferential practices.\footnote{
	For instance, the existence of human races may need to be posited for some eugenicist arguments to succeed;
	or the presupposition of highly-localized specific brain functions
	may be needed for a phrenology-related argument to get through.
	When we intuitively accept the conclusions of arguments we may thereby also commit
	to the existence of their posits (as made explicit in the logical forms of \textit{adequate} formalizations).
	Conversely, such arguments can be attacked by calling into question the mere existence of what they posit.}

\begin{figure}[tp]
	\centering
	\includegraphics[width=\textwidth]{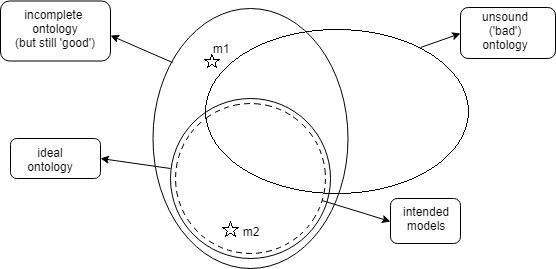}
	\caption{Model ``m1'' is a non-intended model of the `good' yet incomplete ontology;
		whereas ``m2'' is an intended model left out	by the unsound, `bad' ontology.}
	\label{FigIntendedModels}
\end{figure}

Guarino's account leaves some room for partiality (incompleteness) in our representations.
By interpreting natural language, suitably represented as a logical theory,
an interpreter can indeed end up referring to entities other than those originally intended or conceiving them in undesired ways.
In such case we would say that some of the models (interpretations) of the theory are not among its intended models
(see Fig.~\ref{FigIntendedModels}).
In Guarino's view, what is intended becomes prescribed by the target conceptualization.
For instance ---to put it in informal terms--- the conceptualization I commit to could prescribe that all
fishes live in water. However, in one of the possible interpretations (models) of some theory of us (e.g. about pets),
which includes the sentence ``Nemo lives in the bucket'', the term ``Nemo'' may refer to an object falling under
the ``fish'' predicate and the expression ``the bucket'' may refer to some kind of waterless cage.
If, in spite of this rather clumsy interpretation, the theory is still consistent, we would say that the theory somehow
`underperforms': it does not place enough constraints as to 
preclude undesired interpretations, i.e. has models which are not intended.
Underperforming theories (ontologies) in this sense are more the rule than the exception.
Now let us put the previous discussion in formal terms (taken from Guarino et al. \cite{guarino2009ontology}):

\begin{definition}[Extensional relational structure]
    An extensional relational structure, (or a conceptualization according to Genesereth and
    Nilsson \cite{genesereth1987logical}), is a tuple ${\langle}D,R{\rangle}$ where $D$ is a set called the universe of discourse
    and $R$ is a set of (extensional) relations on $D$.
\end{definition}
\begin{definition}[Extensional first-order structure, or model]
	Let $L$ be a first-order logical language with vocabulary $V$ and $S = {\langle}D,R{\rangle}$ an extensional
	relational structure. An extensional first-order structure (also called model for $L$)
	is a tuple $M = {\langle}S, I{\rangle}$, where $I$ (called extensional interpretation function)
	is a total function $I : V \rightarrow D \cup R$ that maps each vocabulary symbol of
	$V$ to either an element of $D$ or an extensional relation belonging to the set $R$.	
\end{definition}

\begin{definition}[Intensional first-order structure, or ontological commitment]
	Let $L$ be a first-order logical language with vocabulary $V$ and $C = {\langle}D,W,\R{\rangle}$
	an intensional relational structure (i.e. a conceptualization). An intensional
	first-order structure (also called ontological commitment) for $L$ is a tuple
	$K = {\langle}C, \I{\rangle}$, where $\I$ (called intensional interpretation function) is a
	total function $\I : V \rightarrow D \cup \R$ that maps each vocabulary symbol of $V$
	to either an element of $D$ or an intensional relation belonging to the set $\R$.
\end{definition}

\begin{definition}[Intended models of a theory w.r.t. a formal conceptualization]
	Let $C = {\langle}D,W,\R{\rangle}$ be a conceptualization, $L$ a first-order logical language
	with vocabulary $V$ and ontological commitment $K = {\langle}C, \I{\rangle}$. A model $M = {\langle}S, I{\rangle}$,
	with $S = {\langle}D,R{\rangle}$, is called an intended model of $L$ according to $K$ iff

	\begin{enumerate}	
		\item For all constant symbols $c \in V$ we have $I(c) = \I(c)$
		\item There exists a world $w \in W$ such that, for each predicate symbol $v \in V$
		there exists an intensional relation $\rho \in \R$ such that $\I(v) = \rho$ and $I(v) = \rho(w)$
	\end{enumerate}

\noindent{The set $I_{K}(L)$ of all models of $L$ that are compatible with $K$ is called the set of intended models of $L$ according to $K$.}
\end{definition}

\begin{definition}[An ontology]
	Let $C$ be a conceptualization, and $L$ a logical language with vocabulary $V$
	and ontological commitment $K$. An ontology $O_{K}$	for $C$ with vocabulary $V$
	and ontological commitment $K$ is a logical theory consisting of a set of formulas of $L$,
	designed so that the set of its models approximates the set $I_{K}(L)$ of
	intended models of $L$ according to $K$ {\rm (see Fig.~\ref{FigIntendedModels}).}
\end{definition}	

Given the definitions above we might conclude, following Guarino, that an ideal ontology is one whose models
exactly coincide (modulo isomorphism) with the intended ones.
Variations to this ideal would entail (with respect to a formal conceptualization) either
(i) incompleteness: the ontology has models that are non-intended, and thus truth in the conceptualization does not
entail validity in the ontology (logical theory);
or (ii) unsoundness: the ontology rules out intended models. This latter situation is the most critical,
since the ontology \textit{qua logical theory} would license inferences that are not valid in the target conceptualization.\footnote{
Guarino further considers factors like language expressivity (richness of logical and non-logical vocabulary)
and scope of the domain of discourse in having a bearing on the degree to which an ontology specifies a conceptualization.}
(See Fig.~\ref{FigIntendedModels} for a diagrammatic illustration.)

Back to the idea of computational hermeneutics, we concede that there are no means of mechanically verifying coherence
with a conceptualization \textit{simpliciter}, the latter being something tacit, fuzzy and unarticulated.
However, we do have the means of deducing and checking consequences drawn from a logical theory,
and this is indeed the reason why computational hermeneutics relies on the use of automated theorem proving.
Regarding formal conceptualizations, we may have the means of mechanically verifying coherence with them
(since they are well-defined mathematical structures), but this would not bring us much further.
As mentioned before, a formal conceptualization is a fictional mathematical object, which might at best serve as a model
(in the sense of being an approximated representation) for some real ---though unarticulated--- conceptualization.
The notion of formal conceptualization has been brought forth in order to enable the previous theoretical analysis
and foster understanding. Hopefully, we can now make better sense of the claim that computational hermeneutics
works by iteratively evolving a logical theory towards adequately approximating a conceptualization, thus making the latter explicit.

\subsection{An Idealized Interpretive Approach} \label{subsec:conceptualizations-ideal}

\begin{figure}[tp]
	\centering
	\includegraphics[width=\textwidth]{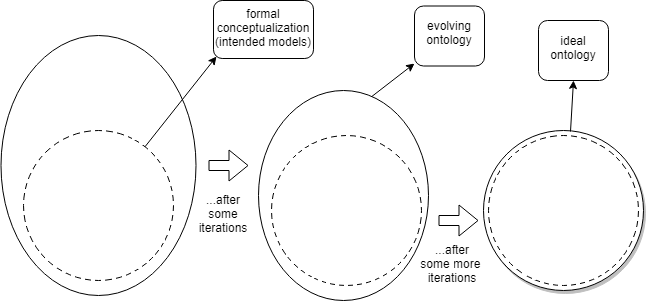}
	\caption{An idealized, straightforward interpretive approach.}
	\label{FigIdealizedApproach}
\end{figure}

As illustrated above, in artificial intelligence and knowledge engineering, ontologies correspond to logical theories
and introduce a vocabulary (logical and non-logical) together with a set of formalized sentences (formulas).
Some of these formulas: axioms, directly aim at ruling out unintended interpretations of the vocabulary so that
as many models as possible correspond to those worlds (i.e. maximal states of affairs) compatible with our target
(formal) conceptualization.
Theorems also do this, but indirectly, by constraining axioms (since the former have to follow from the latter).
Hence, an ontology \textit{qua logical theory} delimits what is possible `from the outside',
i.e. it tells us which interpretations of the given symbols (vocabulary) are not acceptable with respect to
the way we intuitively understand them.
Every time we add a new sentence to the theory we are rejecting those symbols' interpretations (i.e. models) which render the
(now augmented) theory non-valid. Conversely, when removing a sentence some interpretations (models) become acceptable again.
We can thus think of a mechanism to iteratively evolve a logical theory by adding and removing axioms and theorems,
while getting appropriate feedback about the adequacy of our theory changes.
Such a mechanism would converge towards an optimal solution insofar as we can get some kind of quantitative feedback regarding
how well the models of our theory are approximating the intended ones (i.e. the target formal conceptualization).
As discussed more fully below, this is a quite simplistic assumption. However, it is still worth depicting for analysis purposes
as we did for (fictional) formal conceptualizations. This is shown in Fig.~\ref{FigIdealizedApproach}.

\begin{figure}[tp]
	\centering
	\includegraphics[width=\textwidth]{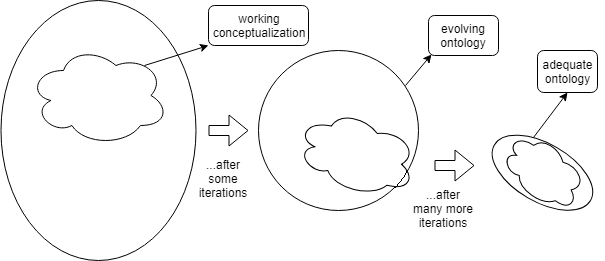}
	\caption{A more realistic view, where a conceptualization is a moving target.}
	\label{FigHermeneuticApproach}
\end{figure}

As mentioned before, \textit{formal} conceptualizations do not exist in real life.
We do not start our interpretive endeavors with a listing of all individual entities in the domain of discourse together
with the relations that may acceptably hold among them.
We have therefore no way to actually determine which are our intended models.
In the situation depicted in Fig.~\ref{FigIdealizedApproach} we were chasing a ghost.
Moreover ---to make things more interesting--- our conceptualization indeed changes as we move through our interpretive process:
We learn more about the concepts involved and about the consequences of our beliefs.
We can even call some of them into question and change our minds.
Fig.~\ref{FigHermeneuticApproach} shows a more realistic view of the problem, where we are chasing a moving target.

\begin{figure}[tp]
	\centering
	\includegraphics[scale=0.8]{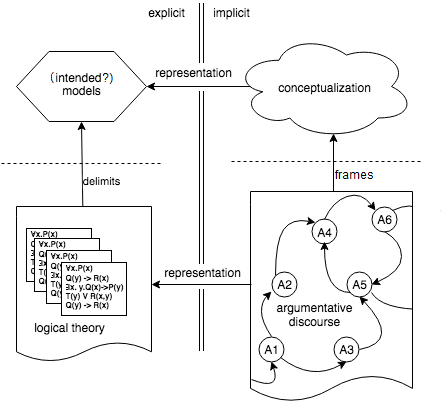}
	\caption{How (models of) logical theories represent conceptualizations.}
	\label{FigRepresentation}
\end{figure}

However, not all hope is lost. In the context of argumentative discourse, we still believe
in the possibility of devising a feedback mechanism which can account for the adequacy of our interpretive logical theories.
Instead of contrasting sets of models with formal conceptualizations,
we will be putting sets of formalized sentences and (extracts from) natural-language arguments side-by-side and then
comparing them among different dimensions. Thus, we will be reversing the account given by Guarino and his colleagues,
by arguing that conceptualizations first originate (i.e. come into existence in an explicit and articulated way)
in the process of developing an interpretive theory. Conceptualizations correspond with `objective reality' to a greater or
lesser extent as they license inferential moves which are endorsed by a linguistic community in the 
context of some argumentative discourse (seen as a network of arguments).
The diagram shown in Fig.~\ref{FigRepresentation} gives a general idea of our interpretive approach.

\section{A Hermeneutic Approach} \label{sec:hermeneutic}

Argumentative discourse and conceptualizations are two sides of a coin. In the previous section we briefly mentioned
the idea of putting logical theories and natural-language arguments side-by-side
and then comparing them among different dimensions. One of those dimensions ---indeed, the main one--- is \textit{truth},
or more appropriately for arguments: \textit{logical correctness}.\footnote{
	Logical correctness encompasses, among others, the more traditional concept of logical validity.
	Our working notion of logical correctness also encompasses axioms/premises consistency and lack of circularity
	(no \textit{petitio principii}) as well as avoiding idle premises. Other accounts may consider different criteria.
	We have restricted ourselves to the ones that can be efficiently computed with today's automated reasoning technology.
	See \cite{peregrin2017reflective} for an interesting discussion of logical (in)correctness.}
This idea has been inspired by Donald Davidson philosophy of language, in particular, his notion of \textit{radical interpretation}.
In this section we aim at showing how the problem of formalizing argumentative discourse, understood as networks of arguments
mutually attacking and supporting each other, relates to the problem of adequately representing a conceptualization.
We share with the supporters of logical expressivism (see e.g. \cite{BrandomMIE,peregrin2014inferentialism}) the view that logical theories
are means to explicitly articulate the rules implicit in discursive practices.

\subsection{Hermeneutic Circle} \label{subsec:hermeneutic-circle}

Previously, we hinted at the possibility of a feedback mechanism that can account for the adequacy
of our interpretive logical theories.
We commented on the impossibility of directly contrasting the models of our theories with the target conceptualization
(determining the theory's intended models), which would have given us a straightforward search path for
the `best' theory (see Fig.~\ref{FigIdealizedApproach}). The main difficulty is that, aside from their complex tacit nature, conceptualizations are also a moving target (as shown in Fig.~\ref{FigHermeneuticApproach}).
Our conceptualizations are work in progress and continually change as we take part in socio-linguistic practices, especially argumentation.
We may thus call them \textit{working} conceptualizations, since they are invariably subject to revision.
This can be seen both at the individual and societal level. For instance, when we as understanding agents become more aware of
(or even learn something new about) our concepts and beliefs. As a community of speakers, through interpretive reconstruction
of public discourse, we can become more self-conscious of our tacit conceptualizations as framed by prevalent discursive practices,
thus enabling their critical assessment and eventual revision.\footnote{Sharing
	a similar background and motivation, computational hermeneutics 
	might support a technological implementation of contemporary approaches for the \textit{revisionary} philosophical analysis of public discourse like ameliorative analysis (in particular, as presented in \cite{novaes2018carnapian}), conceptual ethics \cite{burgess2013conceptual}, and conceptual engineering (e.g. as discussed in \cite{brun2017conceptual}).}

Adhering to the slogan: ``Every formalization is an interpretation'', the logical analysis of arguments becomes essentially
an interpretive endeavor. A characteristic of this endeavor is its recursive, holistic nature.
The logical forms ---as well as the meanings--- of the individual statements comprising an argument (network) are holistically interconnected.
At any point in our interpretive process, the adequacy of the formalization of
some statement (sentence) will hinge on the way other related
sentences have been formalized so far, and thus depend on the current state of our interpretive theory,
i.e. on our \textit{working} conceptualization. Donald Davidson aptly illustrates this situation in the following passage \cite[p.~140]{davidson2001essays}.
\\
\small
\indent{``\ldots much of the interest in logical form comes from an interest in logical geography: to give the logical form of a sentence
is to give its logical location in the totality of sentences, to describe it in a way that explicitly determines what sentences it entails
and what sentences it is entailed by. The location must be given relative to a specific deductive theory;
so logical form itself is relative to a theory.''}
\normalsize
\\
\indent{We thus conceive this hermeneutic interpretive endeavor as a holistic, iterative enterprise. We start with tentative,
simple candidate formalizations of an argument's statements and iteratively use them as stepping stones on the way to the (improved)
formalization of others. The adequacy of any sentence's formalization becomes tested by evaluating the logical correctness
of the formalized arguments (from which it is a component).
Importantly, evaluating logical correctness is a holistic operation that involves all of an argument's constituent sentences.
That is, the result of any formalization's adequacy test becomes dependent not only upon the choice of the particular formula
that is put to the test, but also upon previous formalization choices for its companion sentences in the arguments composing
a discourse (i.e. an argument network).
Moreover, the formalization of any individual sentence depends (by compositionality) on the way its constituent terms have been
explicated (using meaning postulates). Thus, our approach does justice to the inherent circularity of interpretation, where the whole is understood
compositionally on the basis of its parts, while each part can only be understood in the context of the whole.
In the philosophical literature (particularly in \cite{Gadamer}) this recursive nature of interpretation has been termed: the \textit{hermeneutic circle}.}

\subsection{Radical Interpretation} \label{subsec:hermeneutic-radical}

By putting ourselves in the shoes of an interpreter aiming at `translating' some natural-language argument into a formal representation,
we have had recourse to Donald Davidson's philosophical theory of \emph{radical interpretation} \cite{davidson2001inquiries}.
Davidson builds upon Quine's account of \textit{radical translation} \cite{quine1960word}, which is an account of how it is
\textit{possible} for an interpreter to understand someone's words and actions without relying on any prior understanding of them.\footnote{
	In this sense, Davidson has emphatically made clear that he does not aim at showing how humans \textit{actually} interpret
	let alone acquire natural language. This being rather a subject of empirical research (e.g. in cognitive science and linguistics) \cite{davidson1994radical}.
	However, Davidson's philosophical point becomes particularly interesting in artificial intelligence,
	as regards the design of artificial language-capable machines.}
Since the radical interpreter cannot count on a shared language for communication nor the help of a translator, she cannot directly ask for
any kind of explanations. She is thus obliged to conceive interpretation hypotheses (theories) and put them to the test by gathering
`external-world' evidence regarding their validity. She does this by observing the speaker's use of language in context and also by
engaging in some basic dialectical exchange with him/her (e.g. by making utterances while pointing to objects and asking yes--no kind of questions).

Davidson's account of radical interpretation builds upon the idea of taking the concept of \textit{truth} as basic and extracting from it
an account of interpretation satisfying two general requirements: (i) it must reveal the compositional structure of language,
and (ii) it can be assessed using evidence available to the interpreter.
The first requirement (i) is addressed by noting that a theory of truth in Tarski's style \cite{tarski1956concept}
(modified to apply to natural language) can be used as a theory of interpretation.
This implies that, for every sentence \emph{s} of some object language \emph{L},
a sentence of the form: <<``$s$'' is true in $L$ iff $p$>> (aka. T-schema) can be derived, where \emph{p} acts as a translation
of \emph{s} into a sufficiently expressive language used for interpretation (note that in the T-schema the sentence $p$ is
being \emph{used}, while $s$ is only being \emph{mentioned}). Thus, by virtue of the recursive nature of Tarski's definition of
truth, the \emph{compositional} structure of the object-language sentences becomes revealed.
From the point of view of computational hermeneutics, the sentence \emph{s} is to be interpreted in the context of a given argument
(or a network of mutually attacking/supporting arguments). The language \emph{L} thereby corresponds to the idiolect of the speaker
(natural language), and the target language is constituted by formulas of our chosen logic of formalization
(some expressive logic \emph{XY}) plus the meta-logical turnstyle symbol\ \ $\vdash_{\rm XY}$ signifying that an inference
(argument or argument step) is valid in logic \emph{XY}. As an illustration, consider the following instance of the T-schema:
\\
\\
\small
<<``Fishes are necessarily vertebrates'' is true [in English, in the context of argument A] iff $A_1, A_2,..., A_n\ \vdash_{\rm MLS4}$ ``${\forall}x{.}\ \textit{Fish(x)}\  {\rightarrow}\ {\Box}\textit{Vertebrate(x)}$''>>
\normalsize
\\
\\
where $A_1, A_2,..., A_n$ correspond to the formalization of the premises of argument \emph{A} and the turnstyle $\vdash_{\rm MLS4}$
corresponds to the standard logical consequence relation in the chosen logic of formalization, e.g. a modal logic S4 (MLS4).\footnote{
	As will be described in section \ref{subsec:implementation-embeddings}, the semantical embeddings approach \cite{J23,J41}
	allows us to embed different non-classical logics (modal, deontic, intuitionistic, etc.) in higher-order logic (as meta-language),
	and to combine them dynamically by adding and removing axioms.}
This toy example aims at illustrating how the interpretation of a statement relates to its logic of formalization and to the inferential
role it plays in a single argument. Moreover, the same approach can be extended to argument networks. In such cases, instead of using
the notion of logical consequence (represented above as the parameterized logical turnstyle $\vdash_{\rm XY}$), we can work with the
notion of argument \emph{support}. It is indeed possible to parameterize the notions of \emph{support} and \emph{attack} common in
argumentation theory with the logic used for argument's formalization \cite{B18}.

The second general requirement (ii) of Davidson's account of radical interpretation states that the interpreter has access to objective
evidence in order to judge the appropriateness of her interpretation hypotheses, i.e., access to the events and objects in the `external world'
that cause statements to be true ---or, in our case, arguments to be valid.
In our approach, formal logic serves as a common ground for understanding. For instance, computing the logical validity
(or a counterexample) of a formalized argument constitutes the kind of objective ---or more appropriately: intersubjective--- evidence
needed to ground the adequacy of our interpretations, under the \emph{charitable} assumption that the speaker
(i.e. whoever originated the argument) follows, or at least accepts, similar logical rules as we do. In computational hermeneutics,
the computer acts as an (arguably unbiased) arbiter deciding on the correctness of arguments in the context of some encompassing
argumentative discourse ---which itself tacitly frames the conceptualization that we aim at explicitating.

\subsection{The Principle of Charity} \label{subsec:hermeneutic-charity}

A central notion in Davidson's account of radical interpretation is the \emph{principle of charity},
which he holds as a condition for the possibility of engaging in any kind of interpretive endeavor.
The principle of charity has been summarized by Davidson by stating that
``we make maximum sense of the words and thoughts of others when we interpret in a way that optimizes agreement'' \cite[p.~197]{davidson2001inquiries}.
Hence the principle builds upon the possibility of intersubjective agreement about external facts among speaker and interpreter.
The principle of charity can be invoked to make sense of a speaker's ambiguous utterances and,
in our case, to presume (and foster) the correctness of an argument. Consequently, in computational hermeneutics we assume from
the outset that the argument's conclusions indeed follow from its premises and disregard formalizations that do not do justice to
this postulate. Other criteria like avoiding inconsistency and \textit{petitio principii} are also taken into account.
At the argument network level, we foster formalizations which honor the intended dialectic role of arguments,
i.e. attacking or supporting other arguments as intended.

\subsection{Adequacy Criteria of Formalization} \label{subsec:hermeneutic-criteria}

We start to see how the problem of formalizing natural-language arguments relates to the problem of explicitating a conceptualization.
The interpretive theory that gradually evolves in the holistic, iterative computational-hermeneutic process
is composed of those meaning postulates which have become `settled' during the (many) iterations involving arguments' formalization and assessment: It is a logical theory.
Not surprisingly, the quality of our resulting theory will be subordinate to the adequacy of the involved argument formalizations.
Since arguments are composed of statements (sentences), the question thus arises: What makes a certain sentence's formalization better than another?
This topic has indeed been discussed sporadically in the philosophical literature during the last decades
(see e.g. \cite{blau1978dreiwertige,sainsbury1991,epstein1994semantic,brun2004richtige,baumgartner2008adequate}).
More recently, the work of Peregrin and Svoboda \cite{peregrin2013criteria,peregrin2017reflective} has emphasized the role of inferential criteria for assessing the
adequacy of argument formalizations. These criteria are quite in tune, because of their inferential nature, with the holistic picture
of meaning we present here. Furthermore, they lend themselves to mechanization with today's automated reasoning technologies. We recall these inferential criteria below (from \cite{peregrin2017reflective}):

(i) The \emph{principle of reliability}:  
``$\phi$ counts as an adequate formalization of the sentence \emph{S} in the logical system \emph{L} only if the following holds: If an argument form in which $\phi$ occurs as a premise or as the conclusion is valid in \emph{L}, then all its perspicuous natural language instances in which \emph{S} appears as a natural language instance of $\phi$ are intuitively correct arguments.''

(ii) The \emph{principle of ambitiousness}:  
``$\phi$ is the more adequate formalization of the sentence \emph{S} in the logical
system \emph{L}  the more natural language arguments in which \emph{S} occurs as a
premise or as the conclusion, which fall into the intended scope of \emph{L}
and which are intuitively perspicuous and correct, are instances of valid
argument forms of \emph{S} in which $\phi$ appears as the formalization of \emph{S}.'' \cite[pp.~70-71]{peregrin2017reflective}.

The first principle (i) can be seen as a kind of `soundness' criterion: Adequate formalizations 
must not validate intuitively incorrect arguments. The second principle (ii) has an analogous function to that of `completeness' criteria:
We should always aim at the inferentially most \textit{fruitful} formalization, i.e. the one which renders as logically valid (correct) the highest number of intuitively correct arguments.
Peregrin and Svoboda have proposed further adequacy criteria of a more syntactic nature, considering similarity of grammatical structure and simplicity (e.g. number of occurrences of logical symbols).
They are not considered in-depth here. Moreover, many of the `settled' formulas which
constitute our interpretive (logical) theory will have no direct counterpart in the original natural-language
arguments. They are mostly implicit, unstated premises and often of a definitional nature (meaning postulates).
More importantly, Peregrin and Svoboda have framed their approach to logical analysis as a holistic, give-and-take process aiming at reaching
a state of \textit{reflective equilibrium}.\footnote{
	The notion of \emph{reflective equilibrium} has been initially proposed by Nelson Goodman \cite{goodman1983fact} as an account for the justification of the principles of (inductive) logic and has been popularized years later in political philosophy and ethics by John Rawls \cite{rawls2009theory} for the justification of moral principles. In Rawls' account, ``reflective equilibrium'' refers to a state of balance or coherence between a set of general principles and particular judgments (where the latter follow from the former). We arrive at such a state through a deliberative give-and-take process of mutual adjustment between principles and judgments. More recent methodical accounts of reflective equilibrium have been proposed as a justification condition for scientific theories \cite{elgin1999considered} and objectual understanding \cite{baumberger2016dimensions}, and as a methodology for conceptual engineering \cite{brun2017conceptual}.}
Our work on computational hermeneutics can be seen, relatively speaking, as sketching a possible technological implementation
of Peregrin and Svoboda's (and also Brun's \cite{brun2017conceptual}) ideas. We are however careful in invoking the notion of reflective equilibrium for
other than illustrative purposes as we strive towards clearly defined and computationally amenable termination criteria for our interpretive process.\footnote{
	There are ongoing efforts on our part to frame the problem of finding an adequate interpretive theory
	as an optimization problem to be approached by appropriate heuristic methods.}

Not surprisingly, such a holistic approach involves, even for the simplest cases, a search over a vast combinatoric of
candidate formalizations, whose adequacy has to be assessed at least several hundreds of times, particularly if we take the logic of formalization
as an additional degree of freedom (as we do).
An effective evaluation of the above inferential criteria would thus involve automatically computing the logical validity
and consistency of formalized arguments (i.e. proofs). This is the kind of work automated theorem provers and model finders are built for.
The recent improvements in automated reasoning technology (in particular for higher-order logics) constitute the main enabling factor
for computational hermeneutics, which would otherwise remain unfeasible if attempted manually (e.g. carrying out proofs by hand using natural deduction or tableaux calculi). 

\subsection{Computational Hermeneutics} \label{subsec:hermeneutic-computational}

As concerns human-in-the-loop computational hermeneutics, the dialectical exchange between interpreter and speaker depicted above
takes place between humans and interactive proof assistants 
(think of a philosopher-logician seeking to translate `unfamiliar' metaphysical discourse into a `familiar' logical formalism).
At any instant during our iterative, hermeneutic process, we will have an evolving set of meaning postulates and a set of further,
disposable formulas.
We start each new iteration by extending our working interpretive theory with some additional axioms and theorems
corresponding to the candidate formalization of some chosen argument.\footnote{
	Recall that we think of discourses as networks of mutually supporting/attacking arguments.
	Each formalized argument can be seen as a collection of axioms and theorems; the latter being intended to logically follow
	from a combination of the former plus some further axioms of a definitional nature (meaning postulates).
	This view is in tune with prominent structured approaches to argumentation in artificial intelligence (cf. \cite{besnard2008elements,dung2009assumption}).}
In other words, our theory has been temporarily extended with additional axioms and (possible) theorems corresponding to the
formalizations of an argument's premises and conclusion(s) respectively.
This (extended) theory is most likely to \textit{not be} logically correct. It is now our task as (human or machine) interpreters
to come up with the missing axioms that would validate it without rendering it inconsistent or question-begging, among others.
By doing so, we engage in a dialectical exchange with the computer. The questions we ask concern, among others, the logical correctness
(validity, consistency, non-circularity, etc.) of the so extended formal arguments.
Harnessing the latest developments in (higher-order) theorem proving, such questions can get answered automatically in milliseconds.

During the process of securing logically correct formalizations for each of the arguments in the network,
we continuously select among the collected formulas the ones most \textit{adequate} for our purposes.
For this we draw upon the inferential adequacy criteria presented above ---eventually resorting to further discretionary qualitative criteria.
A subset of the formulas will thus qualify for meaning postulates and are kept for future iterations, i.e. they become `settled'.\footnote{
	As mentioned in section \ref{subsec:conceptualizations-ontologies}, there is no definitive criteria for distinguishing meaning postulates from others
	(cf. ontological vs. non-ontological or TBox vs. ABox sentences). The heuristics for labeling sentences as meaning postulates
	thus constitute another degree of freedom in our interpretive process, which we address primarily (but not exclusively) by means of inferential
	adequacy criteria. Moreover, our set of meaning postulates can at some point
	become inconsistent, thus urging us to mark some of them for controlled removal.
	In this aspect, our approach resembles reason-maintenance and belief-revision frameworks (cf. \cite{doyle1992reason}).}
Those meaning postulates can be identified with an ontology as described in section \ref{subsec:conceptualizations-ontologies}.
They introduce semantic constraints aimed at ensuring that any instantiation of the non-logical vocabulary is in line with our working conceptualization
(i.e. with our tacit understanding of the real-world counterparts these symbols are intended to denote).
This is the way argument analysis helps our interpretive theories evolve towards better representing the conceptualization implicit in argumentative discourse.

Removing the human from the loop would, arguably, leave us with an artificial language-capable system.
However, to get there we first need to find a replacement for human ingenuity in the most important `manual' steps currently involved in the human-in-the-loop process:
(i) generating a rough and ready initial formalization for arguments (which we may call ``bootstrapping'');
and (ii) the abductive process of coming up with improved candidate formalizations and implicit premises.
Not surprisingly, these are very difficult tasks; however, they do not belong to science-fiction.
Current developments in natural language processing and machine learning are important enablers for pursuing this ambitious goal.
Another important aspect for both human-in-the-loop and fully-automated computational hermeneutics is the availability of databases of natural-language arguments tagged as either correct or incorrect, as the case may be.
After choosing the area of discourse whose conceptualization we are interested in, we proceed to gather relevant arguments from the relevant
databases and form an argument network with them.\footnote{
	Argument databases and arguments extracted from text sources usually provide information on support and attack relations
	(see \cite{BudzynskaArgumentMining,lippi2016argumentation} and references therein). Another alternative is to dynamically
	construct the needed arguments by using the working theory plus hypothetical premises and conclusions as building stones.
	Those arguments would then be presented, in an interactive way, to the user for rejection or endorsement.
	This mode of operation would correspond to a kind of inverted (we could call it `Socratic') question-answering system.}
Note that, in contrast to highly quantitative, big-data approaches, computational hermeneutics engages in deep semantic analysis of natural-language,
and thus needs far less data to generate interesting results. As previous case studies for human-in-the-loop computational hermeneutics 
applied to ontological arguments have shown (see e.g. \cite{J38,B18} and section~\ref{subsec:implementation-example} below),
working with a handful of sentences already provides interesting characterizations (as meaning postulates) for the
metaphysical concepts involved, e.g. (necessary) existence, contingency, abstractness/concreteness, Godlikeness, essence, dependence, among others.

\section{Examples and Implementation Approaches} \label{sec:implementation}

In this section we summarily discuss previous case studies and 
depict technological solutions for the implementation of computational hermeneutics.
We start by highlighting the essential role of \textit{higher-order} automated theorem proving and the related technique
of \textit{semantical embeddings}. We then depict a landscape of different tools and technologies which are currently
being tested and integrated to pursue an (increasingly) automated computational hermeneutics.

\subsection{Semantical Embeddings} \label{subsec:implementation-embeddings}

Our focus on theorem provers for higher-order logics is motivated by the notion of \textit{logical pluralism}: the view that logics are theories of argument's validity, where
different yet \textit{legitimate} logics may disagree about which argument forms are valid.
In order to pursue a truly logical-pluralist approach in computational hermeneutics, it becomes essential that the (candidate) logics of formalization also form part of our evolving logical theory.
That is, the logic of formalization needs to find its way into the sentences
constituting our interpretive theory. More specifically, the (meta-logical) definitions for the logical vocabulary used to encode
argument's sentences become a further set of axioms which can also be iteratively varied, added and removed as we go.
Thus, computational hermeneutics targets the flexible combination of different kinds of classical and
non-classical logics (modal, temporal, deontic, intuitionistic, etc.) through the technique
of \emph{semantical embeddings} \cite{J23,J41}.

The semantical embeddings approach\footnote{Note that this approach is not related to the similarly named notion of word embeddings in natural language processing (NLP).} harnesses the expressive power of classical higher-order logic (HOL) ---also Church's type theory \cite{J43}--- as a \textit{meta-language} in order to embed the syntax and semantics of another logic
as an object language, thereby turning theorem proving systems for HOL into universal reasoning engines \cite{J41}.
HOL is a logic of functions formulated on top of the simply typed $\lambda$-calculus, which also provides a foundation for functional programming. The semantics of HOL is well understood \cite{J6}.
A variant of this approach, termed \textit{shallow} semantical embeddings (SSE), involves the definition of the logical
vocabulary of a target logic in terms of the non-logical vocabulary (lambda expressions) of our expressive meta-language (HOL).
This technique has been successfully implemented using the interactive proof assistant Isabelle \cite{Isabelle}.

SSE has been exploited in previous work for the evaluation of arguments in philosophy, particularly in metaphysics \cite{C55,J32,C65,J38,B18}.
The formal representation of metaphysical arguments poses a challenging task that requires the combination of different kinds of
expressive higher-order non-classical logics, currently not supported by state-of-the-art automated reasoning technology.
In particular, two of the most well-known higher-order proof assistants: \textit{Isabelle} and \textit{Coq} do not support off-the-shelf reasoning with modal logics. Thus, in order to turn Isabelle into a flexible modal logic reasoner we have adopted the SSE approach.
Isabelle's logic (HOL) supports the encoding of sets via their characteristic functions represented as $\lambda$-terms.
In this sense, HOL comes with a in-built notion of (typed) sets that is exploited in our work for the explicit encoding of the truth-sets that are associated with the formulas of higher-order modal logic.\footnote{
	Note that since Isabelle-specific extensions of HOL (except for prefix polymorphism) are not exploited in our work,
	the technical framework we depict here can easily be transferred to other HOL theorem proving environments.}

\begin{figure}[t]
	\centering
	\includegraphics[scale=0.9]{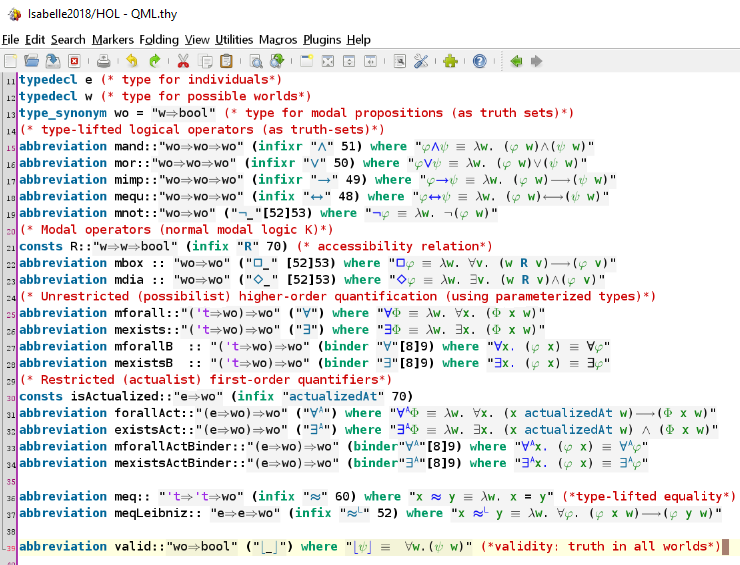}
	\caption{SSE of a higher-order modal logic in Isabelle/HOL.}
	\label{FigSemanticEmbedding}
\end{figure}

As illustrated in Fig.~\ref{FigSemanticEmbedding} (lines 21-23), we can embed a modal logic in HOL by defining the modal $\Box$
and $\Diamond$ operators as meta-logical
predicates in HOL and using quantification over sets of objects of a definite type \emph{w}, representing the type of possible
worlds or world-states. Formula $\Box \varphi$, for example, is modeled as an abbreviation (syntactic sugar)
for the truth-set $\lambda w. \forall v. w~R~v \longrightarrow \varphi v$, where $R$ denotes the accessibility
relation associated with the modal $\Box$ operator. All presented equations exploit the idea that truth-sets in Kripke-style
semantics can be directly encoded as predicates (i.e. sets) in HOL.
Modal formulas $\varphi$ are thus
identified with their corresponding truth sets $\varphi_{w\rightarrow o}$ of predicate type $w\rightarrow o$.
In a similar vein, first-order predicates (i.e. operating on individuals of type $e$) have type $e{\rightarrow}w{\rightarrow}o$.

The semantic embeddings approach gives us two important benefits: (i) we can reuse existing automated theorem
proving technology for HOL and apply it for automated reasoning in non-classical logics
(e.g. free \cite{J40}, modal \cite{J23}, intuitionistic \cite{J21} and deontic \cite{C69} logics);
and (ii) the logic of formalization becomes another degree of freedom in the development of a logical theory
and thus can be fine-tuned dynamically by adding or removing axioms (which are encoded using HOL as a meta-language).

\subsection{Example: Logical Analysis of a Metaphysical Argument} \label{subsec:implementation-example}

\begin{figure}[tp]
	\centering
	\includegraphics[scale=0.85]{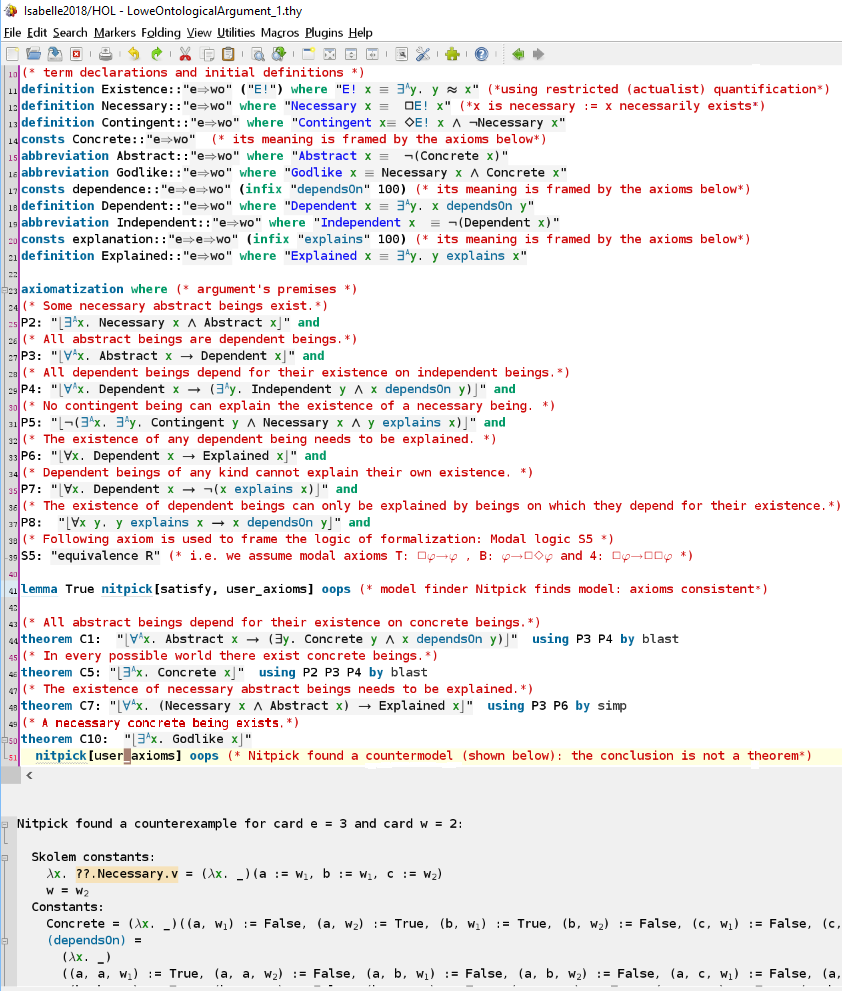}
	\caption{A (non-valid) formalization of Lowe's argument resulting after the first iteration series.
		We can see that the model finder Nitpick \cite{Nitpick} has found a countermodel.
	}
	\label{Lowe1}
\end{figure}

\begin{figure}[t]
	\centering
	\includegraphics[scale=0.9]{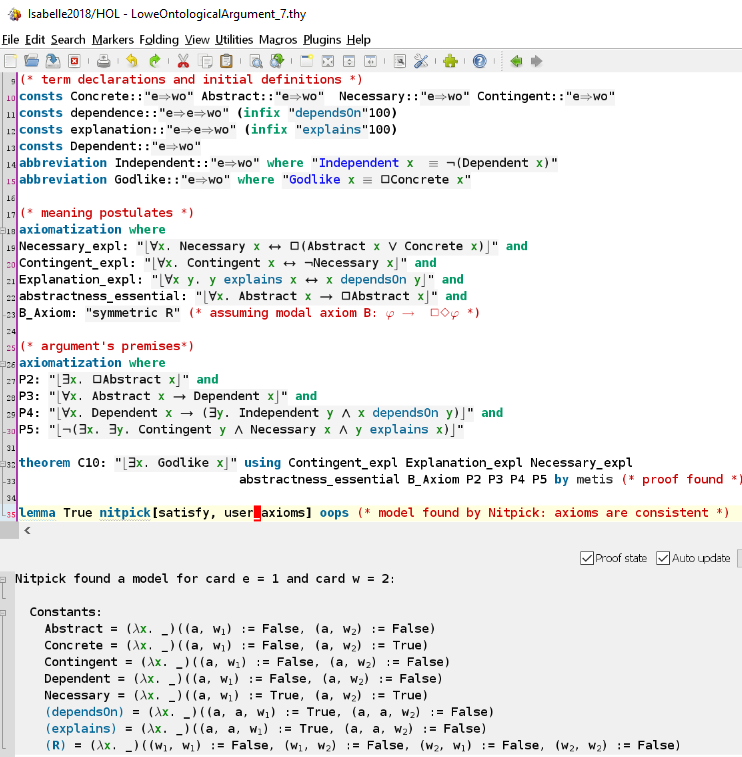}
	\caption{Result after the last iteration series: a satisfying formalization.
		Notice the model generated by Nitpick for the set of axioms (indicating their consistency).}
	\label{Lowe7}
\end{figure}

A first application of computational hermeneutics for the logical analysis of arguments has been presented in \cite{J38} (with its corresponding Isabelle/HOL sources available in \cite{CompHermAFP}).
In that work, a modal variant of the ontological argument for the existence of God, introduced in natural language by the philosopher E.~J.~Lowe \cite{LoweMOA}, has been iteratively analyzed using our computational-hermeneutic approach in a human-in-the-loop fashion and, as a result, a `most' adequate formalization has been found. In each series of iterations (seven in total) Lowe's argument has been formally reconstructed in Isabelle/HOL employing varying sets of premises and logics, and the partial results have been compiled and presented each time as a new variant of the original argument (starting from Fig.~\ref{Lowe1} and ending with Fig.~\ref{Lowe7}). In this fashion, Lowe's argument, as well as our understanding of it, gradually evolves as we experiment with different combinations of definitions, premises and logics for formalization.

Our computational-hermeneutic process starts with an initial, `bootstrapped' formalization. As expected, such a first formalization is often non-well-formed and in most cases logically non-valid; it is generated by manually encoding in an interactive proof assistant (e.g. Isabelle) the author's natural-language formulation using our logically-educated intuition (and with some help from semantic parsers). After this first step, an interpretive give-and-take process is set in motion, and after a few iterations our candidate formalization starts taking shape, as shown in Fig.~\ref{Lowe1} where we have validated some of the argument's partial conclusions, though not the main one (where we get a countermodel from model finder Nitpick\cite{Nitpick} in line 51). During this process we apply our human ingenuity to the abductive task of coming up with new formalization hypotheses (i.e. adding, removing and replacing axioms or definitions) and harness the power of Isabelle's automated reasoning tools to assess their adequacy.

Following the \textit{principle of charity}, our tentative formalizations (logical theories) should become logically valid and meet some inferential (and sometimes syntactical) criteria to our satisfaction. In our case study, this has happened after the seventh iteration series, as shown in Fig.~\ref{Lowe7}, where the now simplified argument\footnote{In \cite{J38} we have produced and discussed other more (or less) complex valid formalizations of this argument before finally settling with the one shown in Fig.~\ref{Lowe7}.} is not only logically valid (this happened already after the second iteration series), but also exhibits other desirable features of a more philosophical nature (as discussed in \cite{J38}); we speak here of arriving at a state of reflective equilibrium (e.g. following \cite{peregrin2017reflective,brun2017conceptual}).

In Fig.~\ref{Lowe7} we can also see how our set of axioms has become split into argument's premises (lines 26-30) and meaning postulates for its featuring concepts (lines 18-23). Here is also shown (line 23) how one of these postulates corresponds to a (meta-logical) axiom constraining the accessibility relation $R$, thus further framing the logic of formalization (modal logic KB). Note that the (meta-logical) definitions shown in Fig.~\ref{FigSemanticEmbedding}, corresponding to the semantical embedding of modal logic $K$ in Isabelle/HOL, can also be seen as further meaning postulates. This embedding has, indeed, also evolved during the interpretive process. For instance, we had to add at some point the formulas that embed restricted (so-called ``actualist'') first-order quantifiers (Fig.~\ref{FigSemanticEmbedding}, lines 30-34), in order to adequately represent the argument (by not trivializing it, as explained in \cite{J38}). As a result of this computational-hermeneutic process, we can see how the meanings of some (rather obscure) metaphysical concepts, like ``necessary existence'', ``contingency'', ``abstractness/concreteness'', ``metaphysical explanation'' and ``dependence'', have become explicitated in the form of logical formulas.

In follow-up work \cite{B18}, we have extended the approach to logical analysis previously sketched in \cite{J38}
by augmenting it with methods as developed in argumentation theory. The idea is now to consider 
the dialectic role an argument plays in some larger area of discourse (represented as a network of arguments).
We have analyzed the contemporary debate around different variants of G\"odel's ontological argument obtaining
some interesting conceptual insights.

\subsection{Technological Feasibility and Implementation Approaches} \label{subsec:implementation-design}

\begin{figure}
	\centering
	\includegraphics[width=\textwidth]{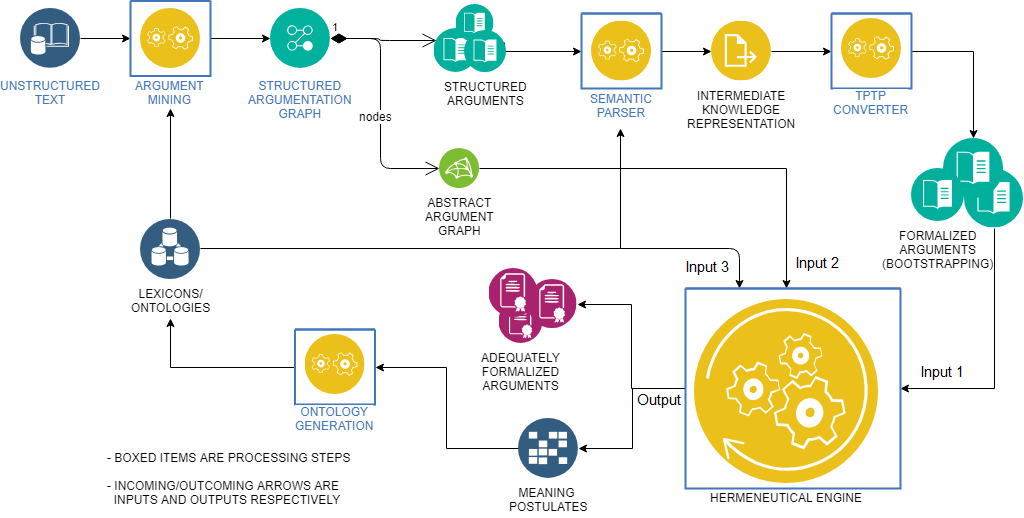}
	\caption{The encompassing technological landscape.}
	\label{FigTechLandscape}
\end{figure}

In this last section we want to briefly discuss some design ideas and challenges
for a software system aimed at automating our computational-hermeneutic approach. These ideas are currently under development.
By sketching a preliminary technological landscape, we argue for the technological feasibility of highly-automated computational hermeneutics in practice. In particular, we want to highlight some of the challenges involved.
In Fig.~\ref{FigTechLandscape}, the main component of the system is the so-called ``hermeneutic engine''
(which will be expanded below in Fig.~\ref{FigHermeneuticEngine}). For the time being let us see it as a black box and discuss its inputs and outputs.

$\bullet$ \textbf{Input 1:} A collection of formalized arguments, i.e. sets of (labeled) formulas.
We bootstrap the interpretive process with some initial rough and ready formalizations encoded in a special format for
use with automated reasoning tools. We consider in particular the TPTP syntax \cite{sutcliffe2017tptp} as it is well supported
in most higher-order theorem provers. In order to arrive at these collections of (bootstrapped) formulas from our
initial natural-language arguments, we rely on argumentation mining technology \cite{BudzynskaArgumentMining,lippi2016argumentation}
and semantic parsers (e.g. the well-known \textit{Boxer} system \cite{bos2008wide}).
Depending on the technological choices, the output format of such tools will most probably be some kind of first-order representation, e.g. Discourse Representation Structures (DRS) (as in \textit{Boxer}). We rely on existing tools supporting the conversion between those first-order representations and TPTP syntax.

$\bullet$ \textbf{Input 2:} Abstract argument network, i.e. a graph whose nodes are the labels of the formalized arguments provided
in \textit{Input 1} and whose edges correspond to attack or support relations.
This kind of structures usually constitute the output of argumentation mining software \cite{BudzynskaArgumentMining,lippi2016argumentation}.

$\bullet$ \textbf{Input 3:} Lexicons and ontologies play an important role in current approaches to semantic parsing and argumentation mining
(e.g. the \textit{Boxer} system has been extended with entity linking to external ontologies \cite{basile2016knews,GangemiEtAl2017}).
Furthermore, existing ontologies can serve as sources during the abductive step of coming up with candidate meaning postulates.

$\bullet$ \textbf{Output:} A collection of (improved) argument formalizations, encoded using appropriate expressive logics.
Importantly, these formalizations have to be logically correct (in contrast to the initial, bootstrapped ones)
and are not restricted to the kind of less-expressive first-order outputs of current semantic parsing technology.
During the process, we maintain a set of sentences common to most formalized arguments and scoring well according to our (mostly inferential) adequacy criteria. These sentences are then labeled as meaning postulates. (Recall from section \ref{subsec:conceptualizations-ontologies} that the heuristic to differentiate between meaning and empirical postulates
is yet another degree of freedom, which is to be determined on the basis of pragmatic considerations.)
In section \ref{subsec:conceptualizations-ontologies} we saw that an ontology (in computer science) can be thought of as a collection of meaning postulates.
Thus, the output of our interpretive process can serve to create new domain-specific ontologies or extend existing ones.

Having discussed the inputs and outputs of our core component, we proceed to roughly
outlining its operation in Fig.~\ref{FigHermeneuticEngine}. 

\begin{figure}
	\centering
	\includegraphics[scale=0.4]{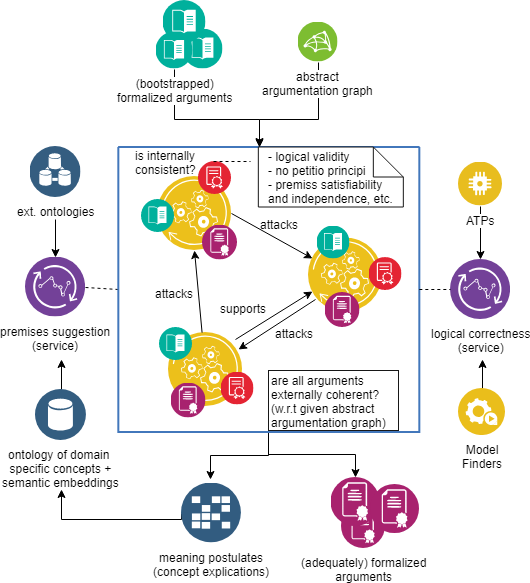}
	\caption{Inside the ``hermeneutic engine''.}
	\label{FigHermeneuticEngine}
\end{figure}

In a nutshell, the engine adds, removes and modifies the arguments' formalized sentences (\textit{Input 1}) so that each argument in the network becomes logically correct, while still fulfilling its intended dialectic role in the network (\textit{Input 2}). Formalized sentences are continuously rated (and picked out) using the adequacy criteria described above.
Computational hermeneutics is thus an iterative and two-layered approach.
In one layer we apply a formalization procedure to each individual argument in parallel,
which consists in temporarily fixing truth-values and inferential relations among its sentences,
and then, after choosing a logic for formalization (i.e. the set of axioms constituting the respective semantical embedding in HOL),
working back and forth on the formalization of its premises and conclusions while getting automatic feedback about the adequacy of the current formalization candidate.
An adequate formalization is one which renders the argument as logically correct and scores high in some additional criteria (syntactic and inferential).
In the next layer we operate at the network level, we continuously choose among the many different combinations of adequate
argument formalizations those which honor their intended dialectic roles, i.e. those which render the arguments as 
successfully attacking or supporting other arguments in the network as intended.
Importantly, we work at these two layers in parallel (or by continuously switching
between them).
In this fashion, by engaging in a methodical `trial-and-error' interaction with the computer,
we work our way towards a proper interpretation of a piece of argumentative discourse
by circular movements between its parts and the whole. As shown in Fig.~\ref{FigHermeneuticEngine}, the output (and usefulness)
of our process consists in generating adequate, logically correct formalizations of 
natural-language arguments, while at the same time articulating logical formulas serving as meaning postulates for those concepts framed by the argumentative discourse in question.
These formulas, following Carnap \cite{carnap1947meaning}, can be seen as conceptual \textit{explications}, thereby fulfilling the task of making the discourse's implicit conceptualization explicit.

\section{Conclusion}
Guarino et al. \cite{guarino1995ontologies,guarino2009ontology} saw in their notion of a (formal) conceptualization a touchstone for evaluating
how good a given logical theory (i.e. an ontology) models or approximates our conception of reality.
As a result of this correspondence with the `real world', inferences drawn using the theory will lead to intuitively correct conclusions.
Drawing on Guarino's exposition and terminology, we have inverted the situation and argued for the thesis that a conceptualization
first becomes articulated by explicitly stating a theory. We can even put it more dramatically and say that a conceptualization first comes into existence
by its being disclosed through an \textit{adequate} logical theory (or more specifically: its models).\footnote{
	Such a reading would be in tune with strong conceptions of existence drawing on the Quinean slogan ``no entity without identity''.}
For the sake of our self-conscious analysis of socio-linguistic practices, we can at any point during an interpretive endeavor
equate a conceptualization with the collection of models of the logical theory under consideration,
provided that this theory satisfies certain adequacy criteria.
This has lead us to discuss what makes an interpretive, logical theory ``adequate'' in this sense. Drawing upon a holistic view of meaning and logical form,
we have approached this issue by considering intersubjective argumentative practices (instead of `objective reality') as our touchstone.
That is, the inferences licensed by our adequate theory are to correspond in a way with the inferences (arguments) endorsed by a community
of speakers (and always in the context of some specific discourse).
We briefly presented Donald Davidson's account of radical interpretation \cite{davidson2001inquiries} together with some
inferential adequacy criteria of formalization recently proposed in the philosophical literature \cite{peregrin2017reflective} and
discussed how they relate to our approach. In the last section we have presented an exemplary application of (human-in-the-loop) computational hermeneutics and sketched some technical implementation approaches towards its further automation.
We showed how current technologies in computational linguistics and automated reasoning enable us
to automate many of the tasks involved in our (currently human-supported) computational-hermeneutic approach. Other, difficult to automate tasks, in particular those 
concerned with abductive reasoning (in higher-order and/or non-classical logics), pose interesting
challenges that we will be addressing in future work. 
There are ongoing efforts to frame our approach in terms of a combinatorial optimization problem.
Given the underdetermined, inexact nature of the problem of logical analysis (and the undecidability of theorem proving in higher-order logics), we do not expect exact algorithms to be found and focus primarily on (highly-parallelizable) metaheuristics.
The integration of machine learning techniques into our approach, in particular regarding the abductive task of suggesting missing (implicit) argument premises, is also being contemplated.

\bibliographystyle{abbrv}

\begin{thebibliography}{10}

\bibitem{basile2016knews}
V.~Basile, E.~Cabrio, and C.~Schon.
\newblock {KNEWS}: Using logical and lexical semantics to extract knowledge
  from natural language.
\newblock In {\em Proceedings of the European Conference on Artificial
  Intelligence (ECAI) 2016 conference}, 2016.

\bibitem{baumberger2016dimensions}
C.~Baumberger and G.~Brun.
\newblock Dimensions of objectual understanding.
\newblock {\em Explaining understanding. New perspectives from epistemology and
  philosophy of science}, pages 165--189, 2016.

\bibitem{baumgartner2008adequate}
M.~Baumgartner and T.~Lampert.
\newblock Adequate formalization.
\newblock {\em Synthese}, 164(1):93--115, 2008.

\bibitem{J41}
C.~Benzm{\"u}ller.
\newblock Universal (meta-)logical reasoning: Recent successes.
\newblock {\em Science of Computer Programming}, 172:48--62, 2019.

\bibitem{J43}
C.~Benzm\"uller and P.~Andrews.
\newblock Church's type theory.
\newblock In E.~N. Zalta, editor, {\em The Stanford Encyclopedia of
  Philosophy}. Metaphysics Research Lab, Stanford University, summer 2019
  edition, 2019.

\bibitem{J6}
C.~Benzm{\"u}ller, C.~Brown, and M.~Kohlhase.
\newblock Higher-order semantics and extensionality.
\newblock {\em Journal of Symbolic Logic}, 69(4):1027--1088, 2004.

\bibitem{C74}
C.~Benzm{\"u}ller and D.~Fuenmayor.
\newblock Can computers help to sharpen our understanding of ontological
  arguments?
\newblock In S.~Gosh, R.~Uppalari, K.~V. Rao, V.~Agarwal, and S.~Sharma,
  editors, {\em Mathematics and Reality, Proceedings of the 11th All India
  Students' Conference on Science \& Spiritual Quest (AISSQ)}, pages 195--226.
  The Bhaktivedanta Institute, Kolkata, 2018.

\bibitem{C69}
C.~Benzm{\"u}ller, X.~Parent, and L.~van~der Torre.
\newblock A deontic logic reasoning infrastructure.
\newblock In F.~Manea, R.~G. Miller, and D.~Nowotka, editors, {\em Proceedings
  of the 14th Conference on Computability in Europe (CiE)}, volume 10936 of
  {\em LNCS}, pages 60--69. Springer, 2018.

\bibitem{J21}
C.~Benzm{\"u}ller and L.~Paulson.
\newblock Multimodal and intuitionistic logics in simple type theory.
\newblock {\em The Logic Journal of the IGPL}, 18(6):881--892, 2010.

\bibitem{J23}
C.~Benzm{\"u}ller and L.~Paulson.
\newblock Quantified multimodal logics in simple type theory.
\newblock {\em Logica Universalis (Special Issue on Multimodal Logics)},
  7(1):7--20, 2013.

\bibitem{J40}
C.~Benzm{\"u}ller and D.~S. Scott.
\newblock Automating free logic in {HOL}, with an experimental application in
  category theory.
\newblock {\em Journal of Automated Reasoning}, 2019.

\bibitem{J32}
C.~Benzm{\"u}ller, L.~Weber, and B.~Woltzenlogel~Paleo.
\newblock Computer-assisted analysis of the {Anderson-H\'{a}jek} controversy.
\newblock {\em Logica Universalis}, 11(1):139--151, 2017.

\bibitem{C55}
C.~Benzm{\"u}ller and B.~Woltzenlogel~Paleo.
\newblock The inconsistency in {G{\"o}del's} ontological argument: A success
  story for {AI} in metaphysics.
\newblock In S.~Kambhampati, editor, {\em IJCAI 2016}, volume 1-3, pages
  936--942. AAAI Press, 2016.

\bibitem{besnard2008elements}
P.~Besnard and A.~Hunter.
\newblock {\em Elements of argumentation}.
\newblock MIT Press, 2008.

\bibitem{Nitpick}
J.~Blanchette and T.~Nipkow.
\newblock Nitpick: A counterexample generator for higher-order logic based on a
  relational model finder.
\newblock In {\em Proc. of ITP 2010}, volume 6172 of {\em LNCS}, pages
  131--146. Springer, 2010.

\bibitem{blau1978dreiwertige}
U.~Blau.
\newblock {\em {Die dreiwertige Logik der Sprache: ihre Syntax, Semantik und
  Anwendung in der Sprachanalyse}}.
\newblock Walter de Gruyter, 1978.

\bibitem{bos2008wide}
J.~Bos.
\newblock Wide-coverage semantic analysis with {Boxer}.
\newblock In {\em Proceedings of the 2008 Conference on Semantics in Text
  Processing}, pages 277--286. Association for Computational Linguistics, 2008.

\bibitem{BrandomMIE}
R.~B. Brandom.
\newblock {\em Making It Explicit: Reasoning, Representing, and Discursive
  Commitment}.
\newblock Harvard University Press, 1994.

\bibitem{brun2004richtige}
G.~Brun.
\newblock {\em {Die richtige Formel: Philosophische Probleme der logischen
  Formalisierung}}.
\newblock Walter de Gruyter, 2004.

\bibitem{brun2017conceptual}
G.~Brun.
\newblock Conceptual re-engineering: from explication to reflective
  equilibrium.
\newblock {\em Synthese}, pages 1--30, 2017.

\bibitem{BudzynskaArgumentMining}
K.~Budzynska and S.~Villata.
\newblock Processing natural language argumentation.
\newblock In P.~Baroni, D.~Gabbay, M.~Giacomin, and L.~van~der Torre, editors,
  {\em Handbook of Formal Argumentation}, pages 577--627. Springer, 2018.

\bibitem{burgess2013conceptual}
A.~Burgess and D.~Plunkett.
\newblock Conceptual ethics {I \& II}.
\newblock {\em Philosophy Compass}, 8(12):1091--1110, 2013.

\bibitem{carnap1947meaning}
R.~Carnap.
\newblock {\em Meaning and necessity: a study in semantics and modal logic}.
\newblock University of Chicago Press, 1947.

\bibitem{carnap1952meaning}
R.~Carnap.
\newblock Meaning postulates.
\newblock {\em Philosophical studies}, 3(5):65--73, 1952.

\bibitem{davidson1994radical}
D.~Davidson.
\newblock Radical interpretation interpreted.
\newblock {\em Philosophical perspectives}, 8:121--128, 1994.

\bibitem{davidson2001essays}
D.~Davidson.
\newblock {\em Essays on actions and events: Philosophical essays}, volume~1.
\newblock Oxford University Press on Demand, 2001.

\bibitem{davidson2001inquiries}
D.~Davidson.
\newblock {\em Inquiries into truth and interpretation: Philosophical essays},
  volume~2.
\newblock Oxford University Press, 2001.

\bibitem{dowty2012introduction}
D.~R. Dowty, R.~Wall, and S.~Peters.
\newblock {\em Introduction to Montague semantics}, volume~11.
\newblock Springer Science \& Business Media, 2012.

\bibitem{doyle1992reason}
J.~Doyle.
\newblock Reason maintenance and belief revision: Foundations vs. coherence
  theories.
\newblock {\em Belief revision}, 29:29--51, 1992.

\bibitem{dung2009assumption}
P.~M. Dung, R.~A. Kowalski, and F.~Toni.
\newblock Assumption-based argumentation.
\newblock In {\em Argumentation in Artificial Intelligence}, pages 199--218.
  Springer, 2009.

\bibitem{elgin1999considered}
C.~Elgin.
\newblock {\em Considered judgment}.
\newblock Princeton University Press, 1999.

\bibitem{epstein1994semantic}
R.~L. Epstein.
\newblock {\em The Semantic Foundations of Logic Volume 2: Predicate Logic},
  volume~2.
\newblock Oxford University Press, 1994.

\bibitem{C65}
D.~Fuenmayor and C.~Benzm{\"u}ller.
\newblock Automating emendations of the ontological argument in intensional
  higher-order modal logic.
\newblock In G.~Kern-Isberner, J.~F{\"u}rnkranz, and M.~Thimm, editors, {\em KI
  2017: Advances in Artificial Intelligence}, volume 10505 of {\em LNAI}, pages
  114--127. Springer, 2017.

\bibitem{CompHermAFP}
D.~Fuenmayor and C.~Benzm\"uller.
\newblock Computer-assisted reconstruction and assessment of {E. J. Lowe's}
  modal ontological argument.
\newblock {\em Archive of Formal Proofs}, Sept. 2017.
\newblock \url{http://isa-afp.org/entries/Lowe_Ontological_Argument.html},
  Formal proof development.

\bibitem{J38}
D.~Fuenmayor and C.~Benzm\"uller.
\newblock A case study on computational hermeneutics: {E.~J.~Lowe's} modal
  ontological argument.
\newblock {\em Journal of Applied Logics -- IfCoLoG Journal of Logics and their
  Applications (special issue on Formal Approaches to the Ontological
  Argument)}, 5(7):1567--1603, 2018.

\bibitem{B18}
D.~Fuenmayor and C.~Benzm{\"u}ller.
\newblock Computational hermeneutics: An integrated approach for the logical
  analysis of natural-language arguments.
\newblock In B.~Liao, T.~Agotnes, and Y.~N. Wang, editors, {\em Dynamics,
  Uncertainty and Reasoning: The Second Chinese Conference on Logic and
  Argumentation}, Logic in Asia series (Studia Logica Library). Springer
  Singapore, 2019.

\bibitem{Gadamer}
H.-G. Gadamer.
\newblock {\em Gesammelte Werke, Bd. 1, Hermeneutik I: Wahrheit und Methode}.
\newblock J.C.B. Mohr (Paul Siebeck), 1960.

\bibitem{GangemiEtAl2017}
A.~Gangemi, V.~Presutti, D.~R. Recupero, A.~G. Nuzzolese, F.~Draicchio, and
  M.~Mongiov\`{i}.
\newblock Semantic web machine reading with {FRED}.
\newblock {\em Semantic Web}, 8(6):873--893, 2017.

\bibitem{genesereth1987logical}
M.~R. Genesereth and N.~J. Nilsson.
\newblock {\em Logical Foundations of Artificial Intelligence}.
\newblock Morgan Kaufmann, 1987.

\bibitem{goodman1983fact}
N.~Goodman.
\newblock {\em Fact, fiction, and forecast}.
\newblock Harvard University Press, 1983.

\bibitem{gruber1993translation}
T.~R. Gruber.
\newblock A translation approach to portable ontology specifications.
\newblock {\em Knowledge acquisition}, 5(2):199--220, 1993.

\bibitem{guarino1995ontologies}
N.~Guarino and P.~Giaretta.
\newblock Ontologies and knowledge bases towards a terminological
  clarification.
\newblock {\em Towards very large knowledge bases: knowledge building \&
  knowledge sharing}, 25(32):307--317, 1995.

\bibitem{guarino2009ontology}
N.~Guarino, D.~Oberle, and S.~Staab.
\newblock What is an ontology?
\newblock In {\em Handbook on ontologies}, pages 1--17. Springer, 2009.

\bibitem{lippi2016argumentation}
M.~Lippi and P.~Torroni.
\newblock Argumentation mining: State of the art and emerging trends.
\newblock {\em ACM Transactions on Internet Technology (TOIT)}, 16(2):10, 2016.

\bibitem{LoweMOA}
E.~J. Lowe.
\newblock A modal version of the ontological argument.
\newblock In J.~P. Moreland, K.~A. Sweis, and C.~V. Meister, editors, {\em
  Debating Christian Theism}, chapter~4, pages 61--71. Oxford University Press,
  2013.

\bibitem{Isabelle}
T.~Nipkow, L.~Paulson, and M.~Wenzel.
\newblock {\em {{Isabelle/HOL}: A Proof Assistant for Higher-Order Logic}},
  volume 2283 of {\em LNCS}.
\newblock Springer, Lecture Notes in Computer Science, 2002.

\bibitem{novaes2018carnapian}
C.~D. Novaes.
\newblock Carnapian explication and ameliorative analysis: a systematic
  comparison.
\newblock {\em Synthese}, pages 1--24, 2018.

\bibitem{peregrin2014inferentialism}
J.~Peregrin.
\newblock {\em Inferentialism: Why rules matter}.
\newblock Springer, 2014.

\bibitem{peregrin2013criteria}
J.~Peregrin and V.~Svoboda.
\newblock Criteria for logical formalization.
\newblock {\em Synthese}, 190(14):2897--2924, 2013.

\bibitem{peregrin2017reflective}
J.~Peregrin and V.~Svoboda.
\newblock {\em Reflective Equilibrium and the Principles of Logical Analysis:
  Understanding the Laws of Logic}.
\newblock Routledge Studies in Contemporary Philosophy. Taylor and Francis,
  2017.

\bibitem{quine1960word}
W.~V.~O. Quine.
\newblock {\em Word and Object}.
\newblock MIT press, 1960.

\bibitem{rawls2009theory}
J.~Rawls.
\newblock {\em A theory of justice}.
\newblock Harvard university press, 2009.

\bibitem{sainsbury1991}
M.~Sainsbury.
\newblock {\em Logical Forms: An Introduction to Philosophical Logic}.
\newblock Blackwell Publishers, 1991.

\bibitem{studer1998knowledge}
R.~Studer, V.~R. Benjamins, and D.~Fensel.
\newblock Knowledge engineering: principles and methods.
\newblock {\em Data \& knowledge engineering}, 25(1-2):161--197, 1998.

\bibitem{sutcliffe2017tptp}
G.~Sutcliffe.
\newblock The {TPTP} problem library and associated infrastructure. {From} {CNF
  to TH0, TPTP v6.4.0}.
\newblock {\em Journal of Automated Reasoning}, 59(4):483--502, 2017.

\bibitem{tarski1956concept}
A.~Tarski.
\newblock The concept of truth in formalized languages.
\newblock {\em Logic, semantics, metamathematics}, 2:152--278, 1956.

\bibitem{uschold1996building}
M.~Uschold.
\newblock Building ontologies: Towards a unified methodology.
\newblock In {\em Proceedings of 16th Annual Conference of the British Computer
  Society Specialists Group on Expert Systems}. Citeseer, 1996.

\end{thebibliography}

\end{document}